\documentclass[conference,a4paper]{IEEEtran}
\IEEEoverridecommandlockouts

\usepackage[hidelinks]{hyperref}
\usepackage[cmex10]{amsmath}
\usepackage{amssymb,amsfonts}
\usepackage{dblfloatfix}

\usepackage[ruled,vlined]{algorithm2e}
\usepackage{graphicx}
\graphicspath{{Figures/PDF/}{Figures/PNG/}}
\usepackage{subcaption}  
\usepackage{float}
\usepackage{booktabs}
\usepackage{siunitx}
\usepackage[numbers,compress]{natbib}
\usepackage{texnames}
\usepackage{bm,bbm}
\usepackage{orcidlink}

\begin{document}

\title{Geometric Flood Depth Estimation: Fusing Transformer-Based Segmentation with Digital Elevation Models
}

\author{	\IEEEauthorblockN{Nhut Le\orcidlink{0009-0002-5223-0505}}
	\IEEEauthorblockA{\textit{Lehigh University}\\
		Bethlehem, PA 18015, US\\
		nhl224@lehigh.edu}
        \and
	\IEEEauthorblockN{Ehsan Karimi\orcidlink{0000-0001-8495-9013}}
	\IEEEauthorblockA{\textit{Lehigh University}\\
		Bethlehem, PA 18015, USA\\
		ehk224@lehigh.edu}
	\and
	\IEEEauthorblockN{Maryam Rahnemoonfar\textsuperscript{*}\orcidlink{0000-0001-9358-2836}}
	\IEEEauthorblockA{\textit{Lehigh University}\\
		Bethlehem, PA 18015, USA\\
		maryam@lehigh.edu \thanks{*Corresponding author}
}}

\maketitle
\begin{abstract}
Post-disaster situational awareness relies heavily on understanding both the extent and the volume of floodwaters. While 2D semantic segmentation provides accurate flood masking, it lacks the vertical dimension required to assess navigability and structural risk. This paper presents a geometric "Water Surface Elevation" approach for estimating flood depth from monocular aerial imagery. Our pipeline utilizes Mask2Former, a state-of-the-art transformer-based segmentation model, to generate precise 2D flood masks. These masks are fused with Digital Elevation Models (DEMs) to identify the water-land boundary, calculate a global water surface elevation ($Z_{water}$), and compute per-pixel depth based on the principle of local hydrostatic equilibrium. We evaluate this workflow using the FloodNet and CRASAR-U-DROIDS datasets, demonstrating how high-performance segmentation can be leveraged to extract 3D volumetric data from 2D imagery without the latency of hydrodynamic simulations.
\end{abstract}


\section{Introduction}
Rapid response to flooding requires more than knowing where water is; responders need to know how deep it is. Deep learning has revolutionized flood extent detection through semantic segmentation, yet extracting volumetric depth from monocular aerial imagery remains a persistent challenge.

\begin{figure}[!t]
\centering
\includegraphics[width=0.8\linewidth]{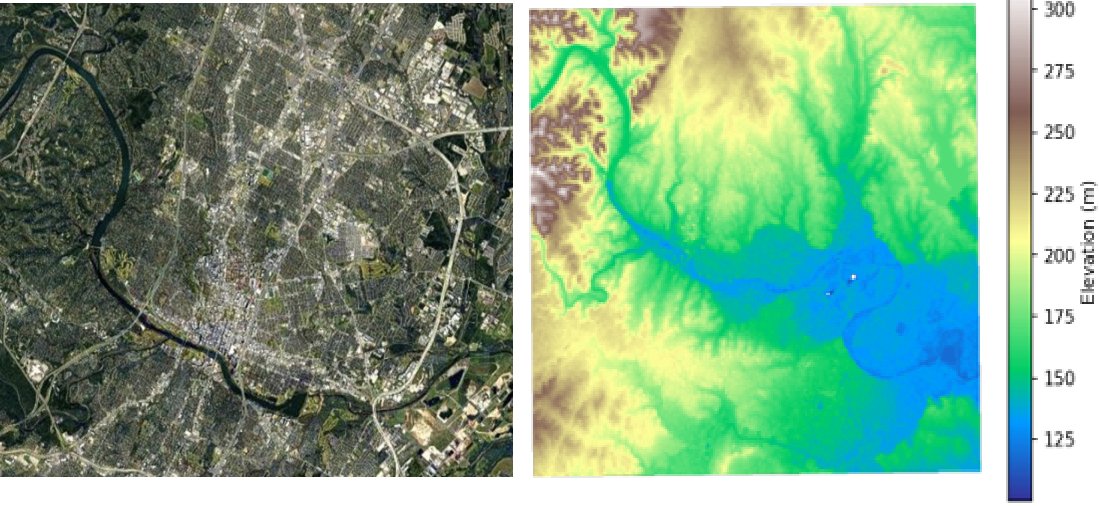}
\caption{\textbf{Data alignment.} \textbf{Left:} Satellite imagery of Austin, Texas. \textbf{Right:} Corresponding DEM aligned to the imagery for terrain height extraction.}
\label{fig:data_alignment}
\end{figure}

Hydrodynamic solvers \cite{Mungkasi_2013, brunner1995hec, hervouet2007hydrodynamics}, which resolve the Shallow Water Equations (SWE), offer physical rigor but suffer from two bottlenecks: latency and data scarcity. In our preliminary experiments, simulating a single 12-hour flood event on a $100 \times 100$ grid required 1-2 days of CPU time. These models also depend on real-time boundary conditions - such as inflow rates and rainfall intensity - that are rarely available in the chaotic aftermath of a disaster. Post-event reference products such as FEMA flood depth rasters \cite{fema2017harvey}, derived from high-water mark surveys and hydraulic modeling, provide accurate ground truth but are typically released weeks to months after an event, making them unsuitable for immediate response.

Machine learning (ML)-based surrogate models reduce inference time but often introduce smoothing artifacts or hallucinate floods in physically impossible locations. We propose a Geometric Depth Estimation pipeline that bypasses both the computational cost of simulation and the instability of data-driven surrogates. By fusing high-precision 2D segmentation masks with pre-existing DEMs, we infer depth geometrically. Operating on the ``bathtub principle,'' our method assumes standing floodwater creates a locally flat surface (see Fig.~\ref{fig:method_schematic}), offering a computationally efficient and physically grounded approach requiring only aerial imagery and open-source elevation data.

While prior geometric approaches \cite{soria2023extent, soria2022sentinel} rely on traditional ML classifiers or SAR-derived masks with coarser boundaries, our contribution is the integration of a transformer network, whose attention mechanism yields sharper waterline delineation - the critical bottleneck in geometric depth accuracy. We also introduce a ``Strict Dryland Contact'' boundary logic with edge-artifact rejection, not present in prior bathtub methods. We utilize FloodNet \cite{rahnemoonfar2021floodnet} to train our segmentation architecture (Mask2Former \cite{cheng2021mask2former}), CRASAR-U-DROIDS \cite{manzini2024crasar} for deployment, and FEMA depth rasters \cite{fema2017harvey} as quantitative ground truth for volumetric validation.

\section{Related Work}
\subsection{Semantic Segmentation in Disaster Scenarios}
Precise segmentation in post-disaster scenarios is essential for distinguishing floodwaters from debris and infrastructure, motivating high-resolution benchmarks like FloodNet \cite{rahnemoonfar2021floodnet} and RescueNet \cite{rahnemoonfar2023rescuenet}. CNNs \cite{paszke2016enet,chen2018encoder} rely on fixed receptive fields, often producing coarse boundary delineation where thin structures such as flooded roads are lost or merged with adjacent water bodies. Transformer-based architectures \cite{cheng2021mask2former,xie2021segformer,oktay2018attention} address this by modeling global context dynamically, enabling sharper class separation. We benchmark these architectures on FloodNet \cite{rahnemoonfar2021floodnet} to select the optimal backbone for the boundary precision required by our depth pipeline.

\subsection{Flood Depth Estimation Strategies}
Depth estimation falls into three paradigms. \textit{Hydrodynamic modeling} (e.g., ANUGA \cite{Mungkasi_2013}, HEC-RAS \cite{brunner1995hec}) offers physical rigor but is computationally prohibitive for real-time response. \textit{Reference-object methods} \cite{song2021automated, liu2023novel} infer depth from visual gauges but fail in aerial imagery where vertical cues are occluded. \textit{Geometric integration} \cite{soria2023extent, soria2022sentinel, 9554975} intersects 2D flood masks with topographic data; our work advances this paradigm by replacing standard classifiers with Mask2Former for high-fidelity waterline identification.
\section{Methodology}
\begin{figure}[!t]
\centering
\includegraphics[width=0.8\linewidth]{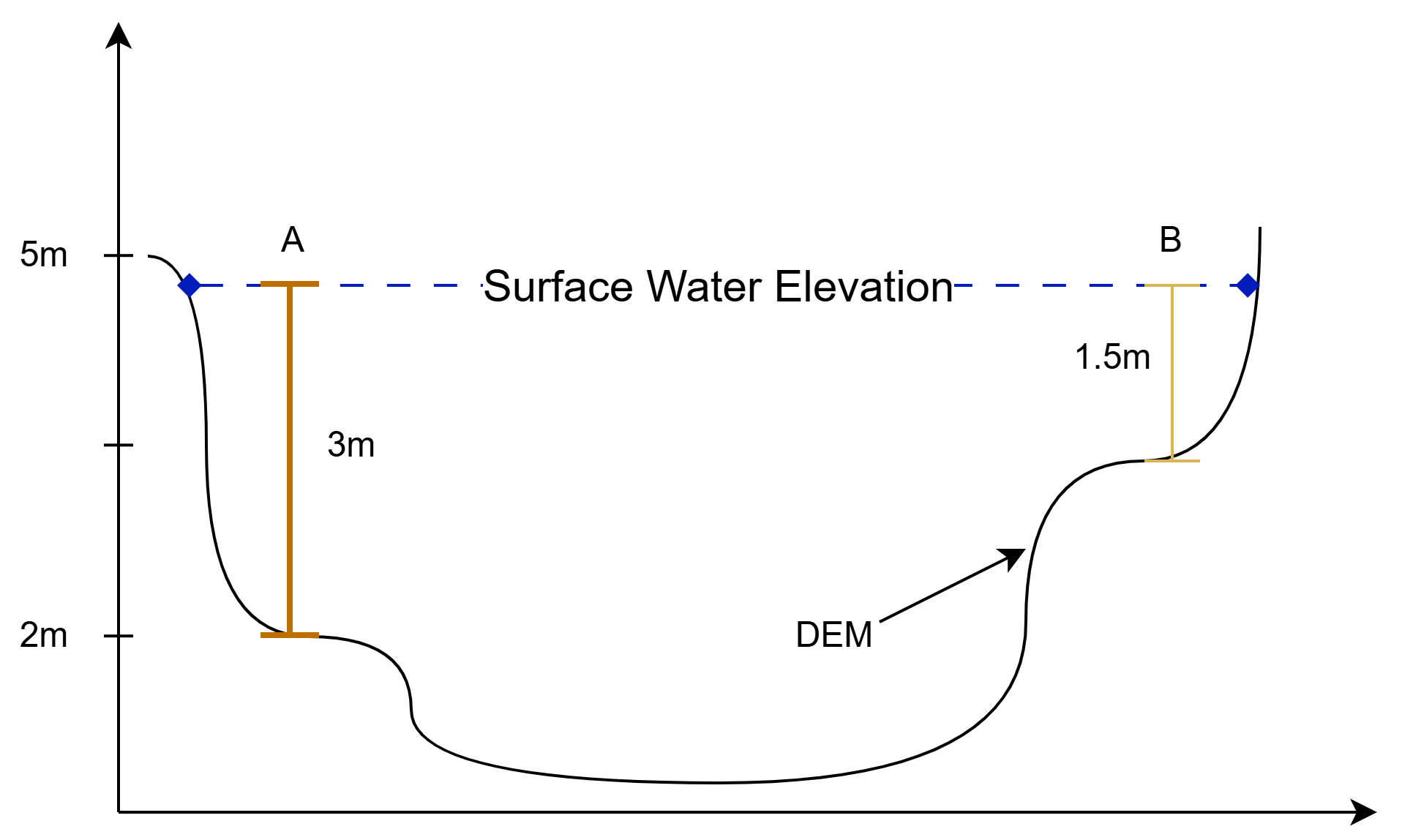}
\caption{Schematic of the Geometric Depth Estimation approach. The pipeline identifies the Water Surface Elevation ($Z_{water}$) by sampling DEM values at the boundary where water meets land. The flood depth at any interior point is then computed as the vertical difference between this global water surface and the local terrain height ($Z_{water} - \text{DEM}$) (points A and B), assuming local hydrostatic equilibrium.}
\label{fig:method_schematic}
\end{figure}
To address the challenge of rapid depth estimation, we propose a geometric approach that relies on direct observation. By fusing high-precision 2D segmentation masks with publicly available topographic data, we bypass the computational costs of hydrodynamic simulation. Our pipeline consists of three stages: Data Acquisition and Preprocessing, Flood Segmentation, and Geometric Depth Calculation.

\subsection{Data Acquisition and Preprocessing}

\subsubsection{Datasets}
We utilize two distinct datasets to ensure robust performance. FloodNet \cite{rahnemoonfar2021floodnet}, which contains high-resolution UAV imagery annotated with fine-grained classes (e.g., "Flooded Building", "Flooded Road"), is used to train our segmentation model. For volumetric validation, we utilize the CRASAR-U-DROIDS \cite{manzini2024crasar} dataset. Unlike standard computer vision benchmarks, CRASAR provides imagery in GeoTIFF format, preserving the critical geospatial metadata required to align pixel coordinates with real-world topography.
\subsubsection{Topographic Data (DEM)}
For each scene in the CRASAR dataset, we programmatically fetch the corresponding Digital Elevation Model (DEM) from the OpenTopography \cite{krishnan2011opentopography} (see Fig. \ref{fig:data_alignment}). To ensure pixel-wise alignment with the high-resolution drone imagery, the fetched DEMs are reprojected and upsampled using bilinear interpolation to match the Coordinate Reference System (CRS) and spatial resolution of the source UAV images. 

\subsubsection{Tiling and Event Filtering}
Since the source CRASAR GeoTIFFs are large-scale orthomosaics exceeding standard model input dimensions, we implement a targeted tiling strategy: (1)~\textbf{Building-Centric Cropping:} $1024\times1024$ tiles are centered on building polygon centroids, ensuring every tile contains structural elements. (2)~\textbf{Hurricane Filtering:} Tiles are restricted to major hurricane events (Harvey, Michael, Ida, Ian, Laura) to ensure floodwater presence. (3)~\textbf{Quality Control:} Tiles with void pixels from irregular orthomosaic borders are automatically discarded.

\subsection{Stage 1: 2D Flood Segmentation}
\subsubsection{Model Selection}
To select the optimal architecture for this task, we fine-tuned three state-of-the-art deep learning models on the FloodNet dataset \cite{rahnemoonfar2021floodnet}: Attention U-Net \cite{oktay2018attention}, SegFormer \cite{xie2021segformer}, and Mask2Former \cite{cheng2021mask2former}. All models were trained on the labeled training split, and quantitative evaluations were performed strictly on the held-out test set to ensure robust generalization. The candidate architectures included:
\begin{itemize}
    \item \textbf{Attention U-Net \cite{oktay2018attention}:} A convolutional network utilizing attention gates in skip connections to highlight salient features.
    \item \textbf{SegFormer \cite{xie2021segformer}:} A hierarchical transformer framework with a lightweight Multi-Layer Perceptron (MLP) decoder.
    \item \textbf{Mask2Former \cite{cheng2021mask2former}:} A mask classification architecture utilizing a transformer decoder with masked attention.
\end{itemize}
\begin{table*}[t]
\centering
\caption{Semantic segmentation performance (mIoU) on the FloodNet dataset. The best results for each category are highlighted in \textbf{bold}.}
\label{tab:segmentation_results}
\resizebox{\textwidth}{!}{%
\begin{tabular}{|l|c|c|c|c|c|c|c|c|c|c|}
\hline 
\textbf{Methods} & \textbf{Mean} & \textbf{\shortstack{Bldg\\(Flooded)}} & \textbf{\shortstack{Bldg\\
(Non-Flooded)}} & \textbf{\shortstack{Road\\(Flooded)}} & \textbf{\shortstack{Road\\(Non-Flooded)}} & \textbf{Water} & \textbf{Tree} & \textbf{Vehicle} & \textbf{Pool} & \textbf{Grass} \\
\hline
Attention-Unet & 69.7 & 58.4 & 76.9 & 47.8 & 83.0 & 74.5 & 81.0 & 53.6 & 63.7 & 88.4 \\
\hline
Segformer & 70.4 & \textbf{63.3} & 79.5 & 51.4 & 83.2 & 73.6 & 82.4 & 46.2 & 57.6 & 88.8 \\
\hline  
Mask2Former & \textbf{73.6} & 62.6 & \textbf{80.4} & \textbf{58.0} & \textbf{84.6} & \textbf{77.3} & \textbf{83.5} & \textbf{59.6} & \textbf{66.7} & \textbf{89.6} \\
\hline
\end{tabular}%
}
\end{table*}

\subsubsection{Inference}
Our segmentation model utilizes a transformer-based architecture pre-trained on the FloodNet dataset \cite{rahnemoonfar2021floodnet}. To adapt the output on CRASAR tiles, we aggregate the original classes into a simplified semantic mask (see Fig. \ref{fig:qualitative_results}(2)). We combine Water and Flooded Roads into a single Flood class, and Building-Flooded and Building-Non-Flooded into a Buildings class; all other categories are treated as Background. This aggregation is essential for urban scenarios, where flooded streets typically constitute the primary connected water body.

\subsection{Stage 2: Geometric Depth Estimation}
We calculate depth based on the "bathtub principle" - assuming that within a connected basin, standing water maintains a locally flat surface level.

\subsubsection{Step 1: Connected Component Analysis (CCA)}
Global depth calculation is often erroneous because different water bodies in a scene (e.g., a flooded road vs. a pool) exist at different elevations. To address this, we apply Connected Component Analysis to treat disjoint water bodies as independent physical systems.
\begin{itemize}
    \item \textbf{Input:} The binary flood mask $M_{binary}$ where $1=$ Water/Flooded Road and $0=$ Otherwise.
    \item \textbf{Operation:} We apply an 8-connectivity labeling algorithm to $M_{binary}$. This scans the mask and assigns a unique ID $k$ to every distinct cluster of connected pixels.
    \item \textbf{Output:} A labeled map $L$ where $L(x,y) = k$ implies pixel $(x,y)$ belongs to the $k$-th independent water body.
\end{itemize}
This ensures that the water level $Z_{water}^{(k)}$ calculated for a specific flood basin $k$ does not erroneously influence the depth calculation of a separate, unconnected body.

\subsubsection{Step 2: Waterline Identification} 
\begin{figure*}[!t]
    \centering
    \setlength{\tabcolsep}{1pt}  
    \renewcommand{\arraystretch}{0.5} 
    \begin{tabular}{ccccc} 
        \small{Input RGB} & \small{Segmentation} & \small{Water Boundary} & \small{DEM} & \small{Est. Depth} \\
        
        \includegraphics[width=0.17\linewidth]{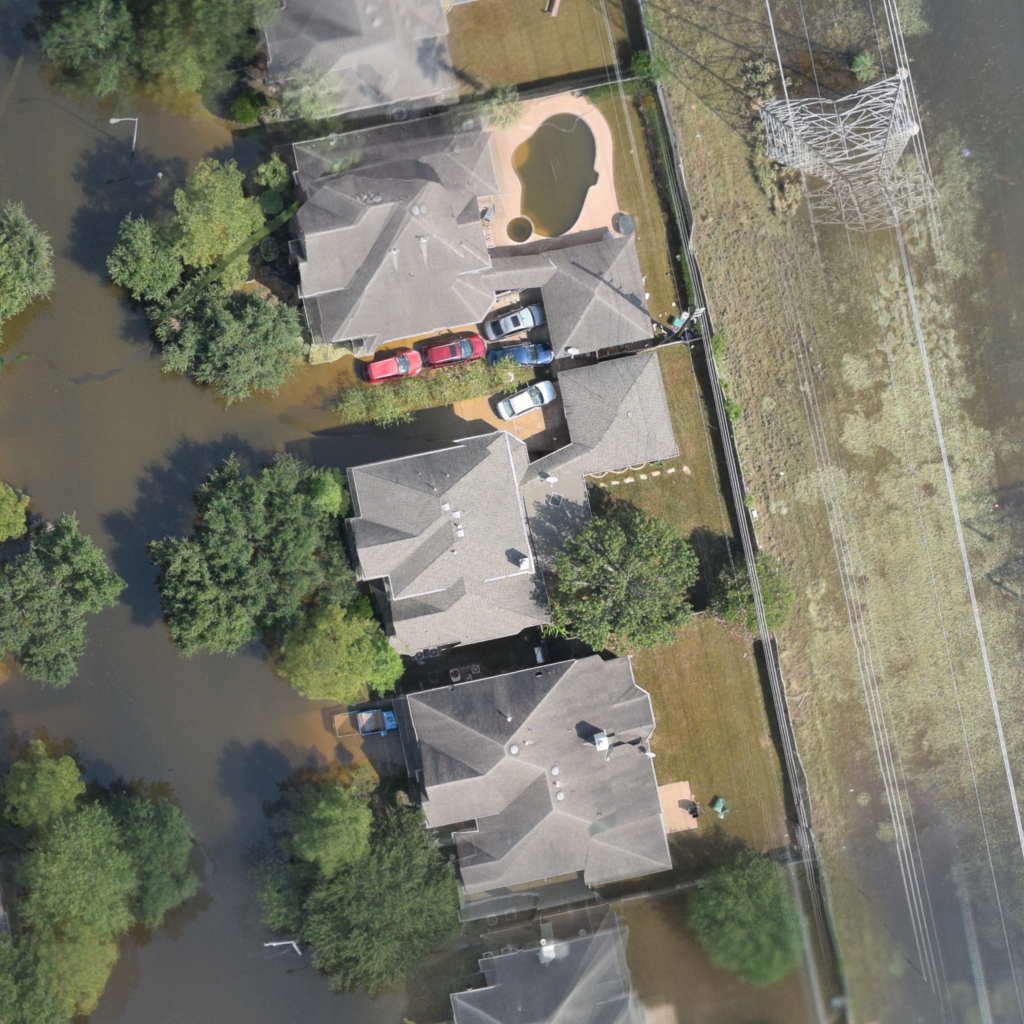} &
        \includegraphics[width=0.17\linewidth]{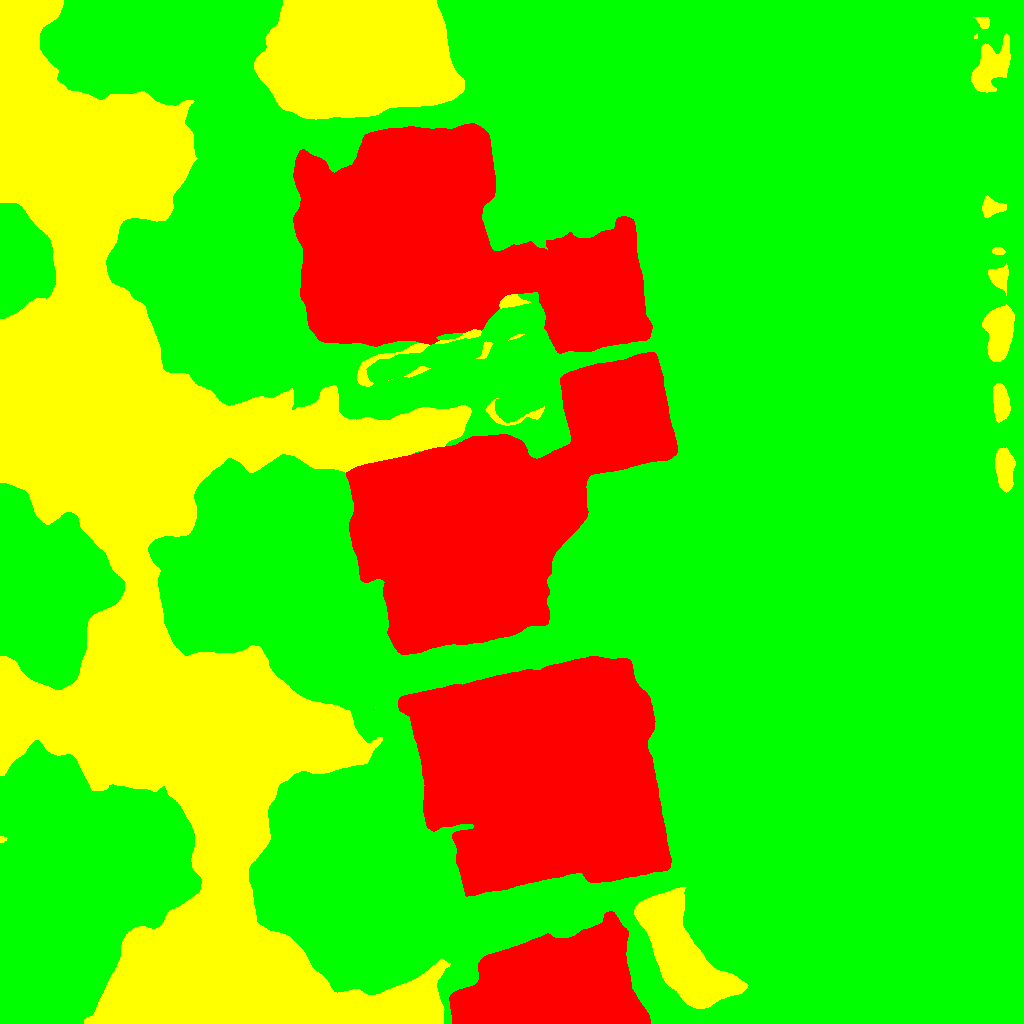} &
        \includegraphics[width=0.17\linewidth]{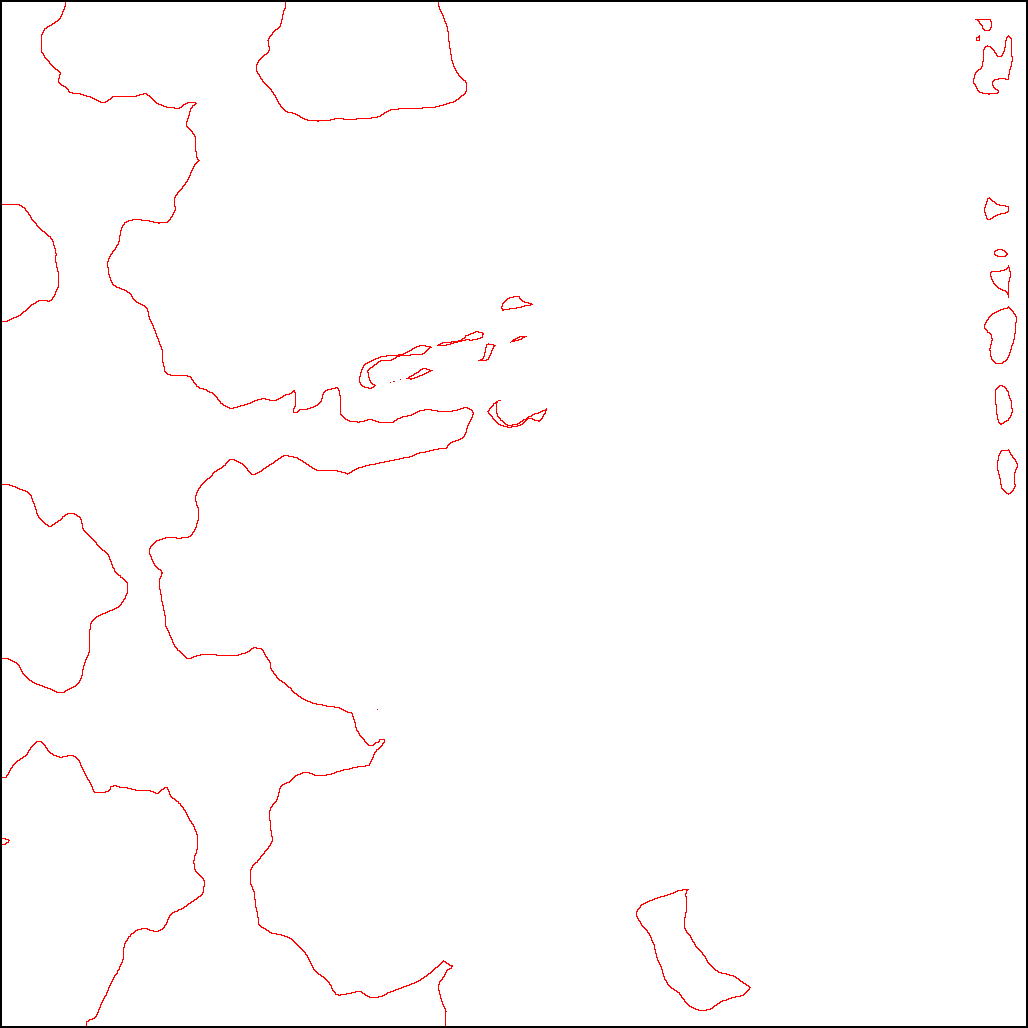} &
        \includegraphics[width=0.17\linewidth]{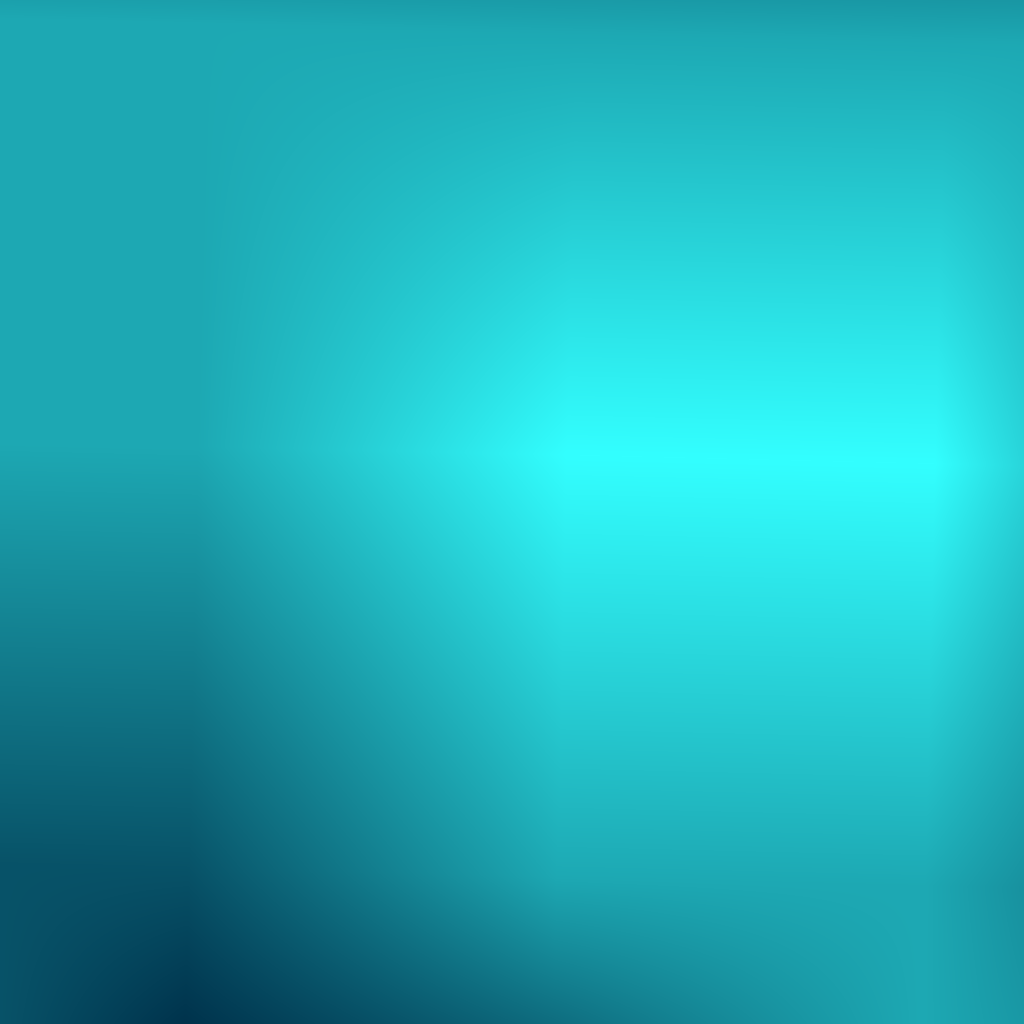} &
        \includegraphics[width=0.2\linewidth]{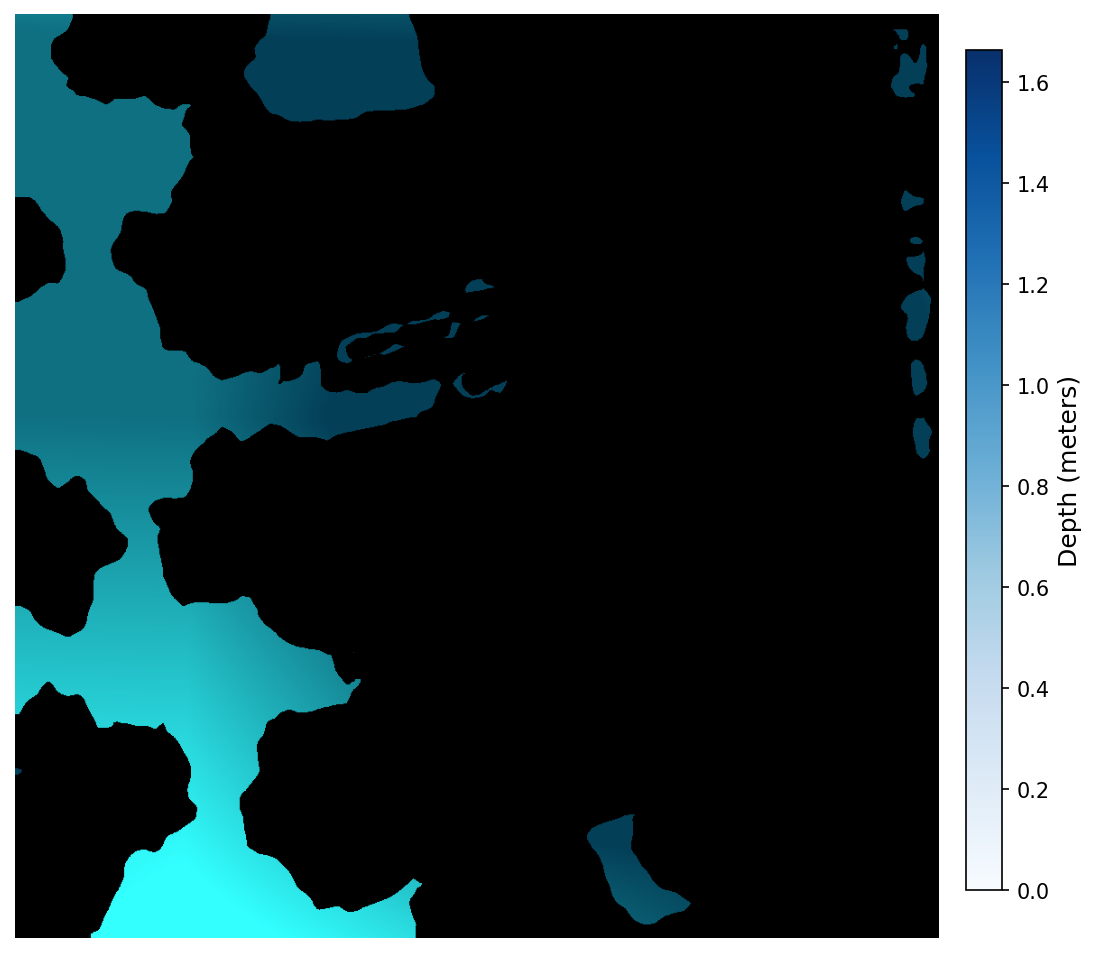} \\
        
     \includegraphics[width=0.17\linewidth]{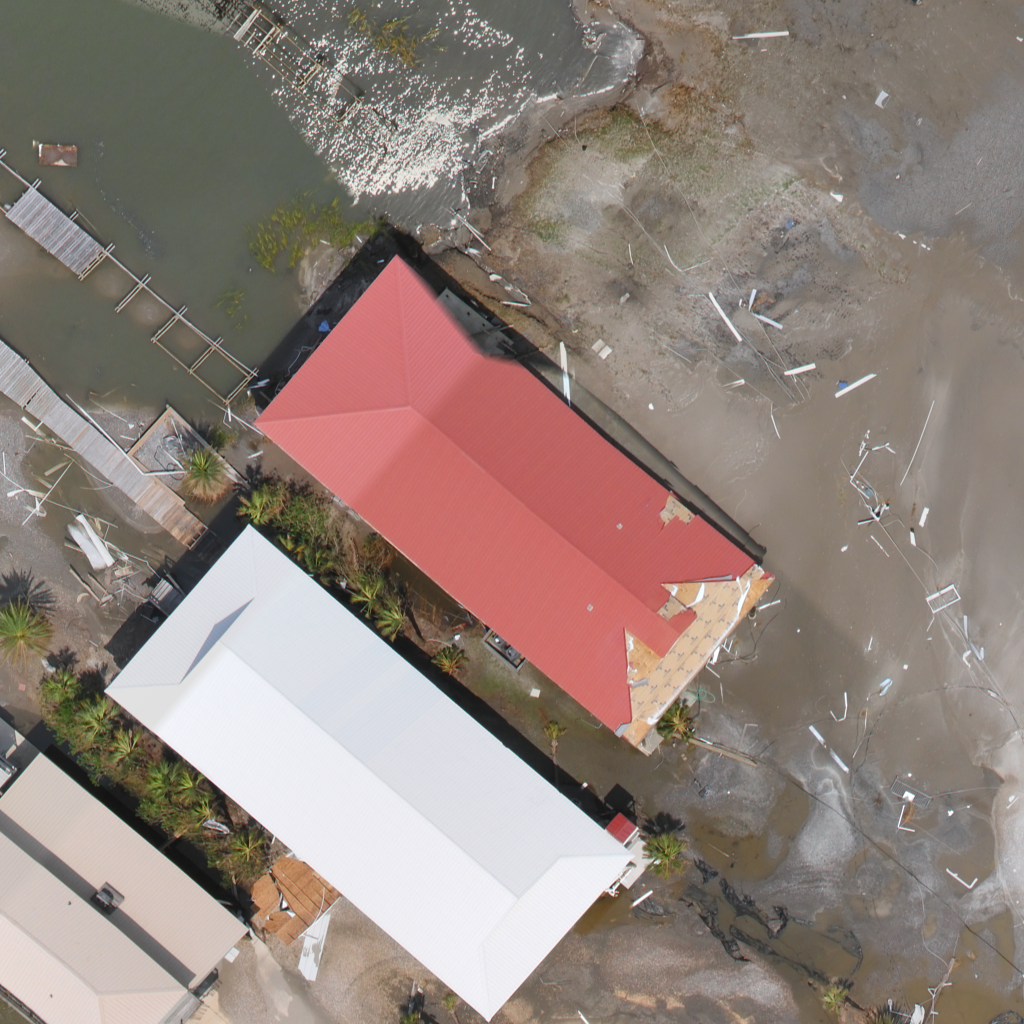} &
        \includegraphics[width=0.17\linewidth]{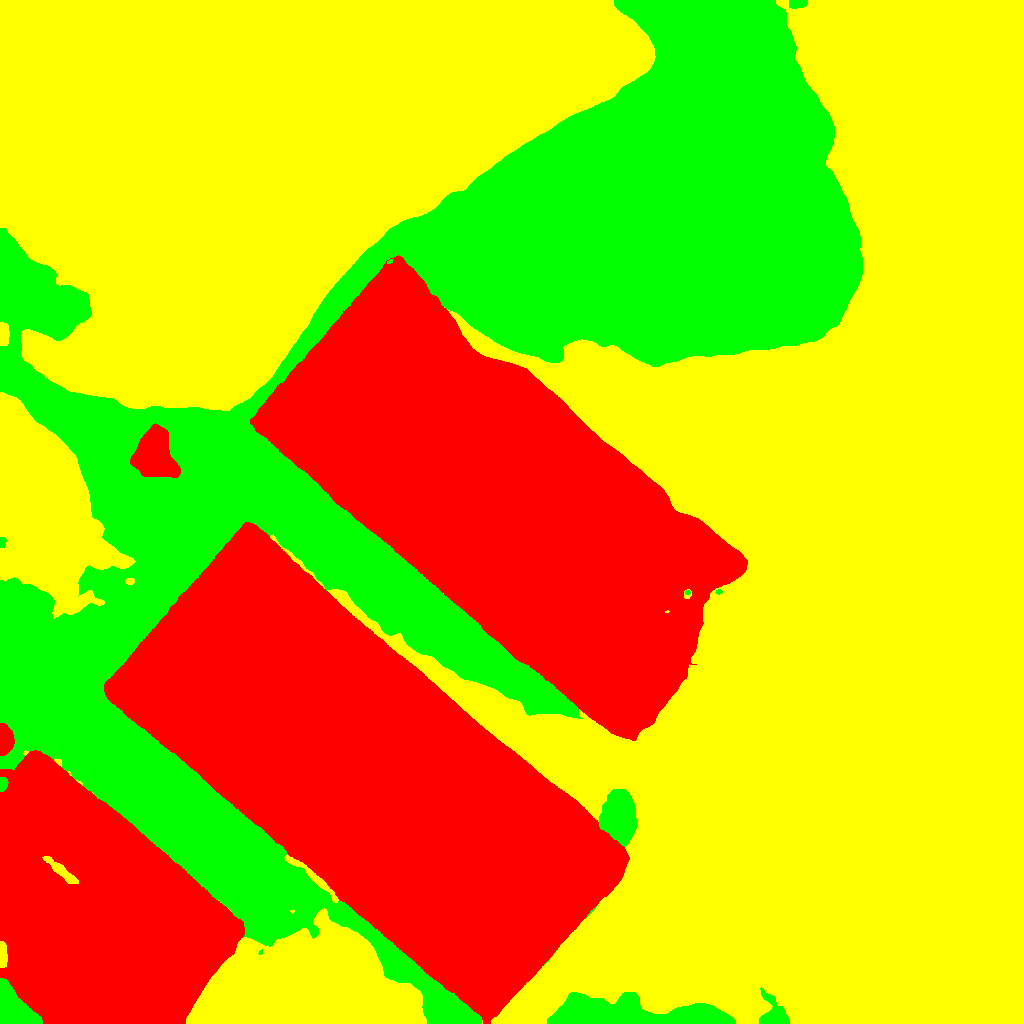} &
        \includegraphics[width=0.17\linewidth]{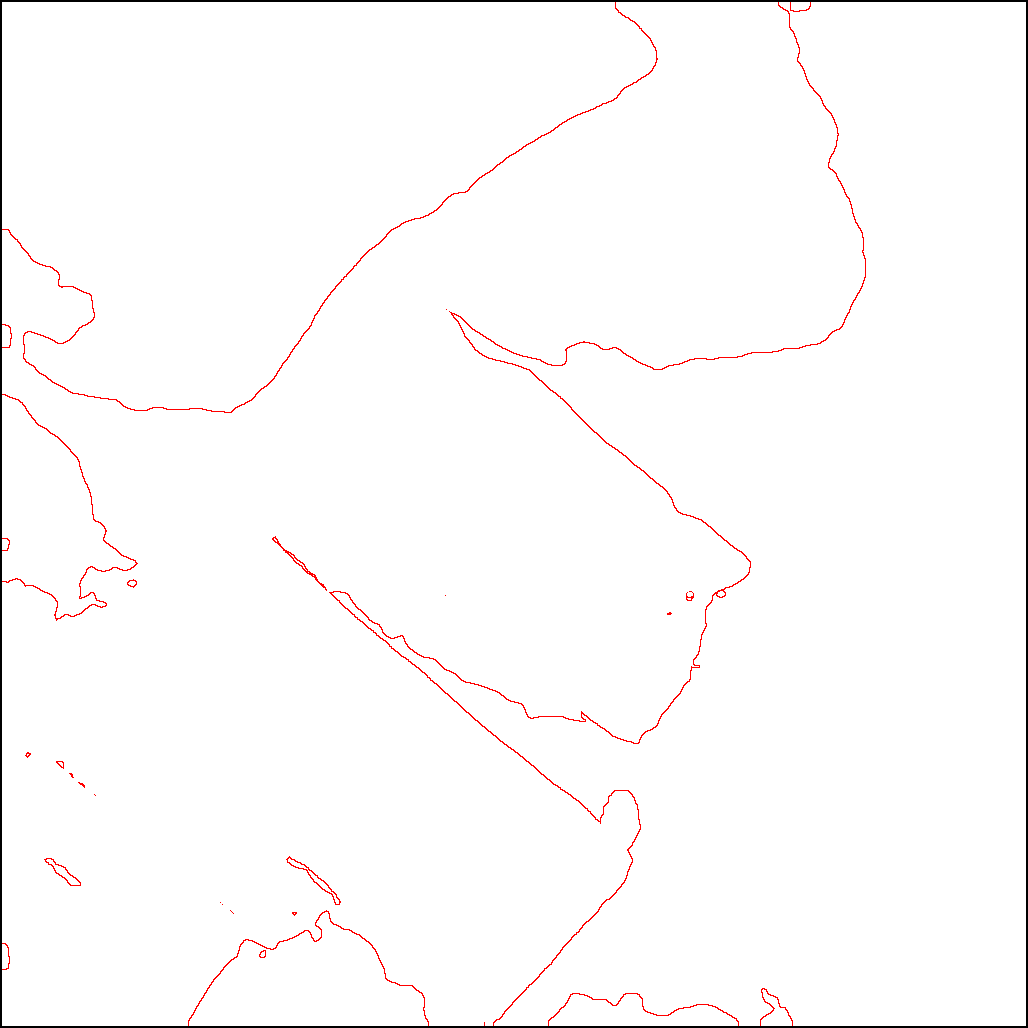} &
        \includegraphics[width=0.17\linewidth]{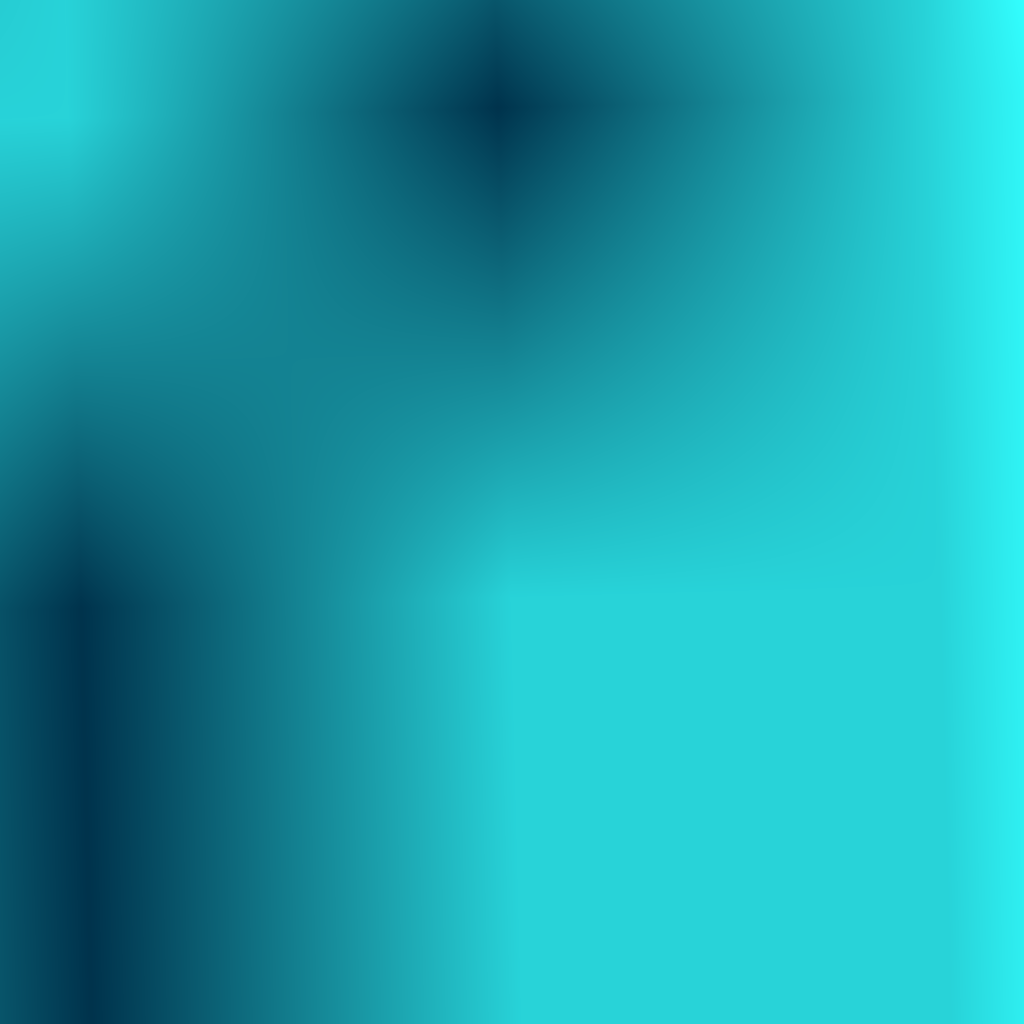} &
        \includegraphics[width=0.2\linewidth]{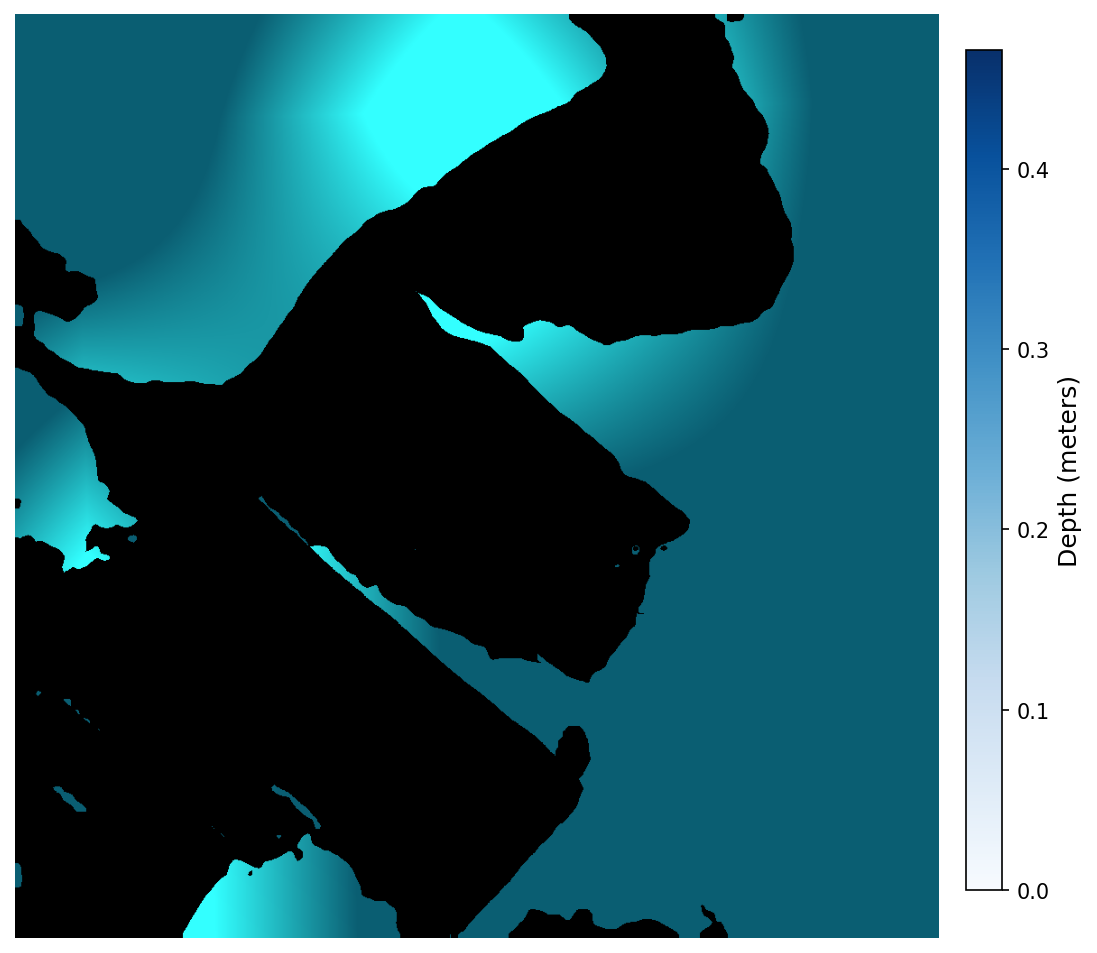} \\

       \includegraphics[width=0.17\linewidth]{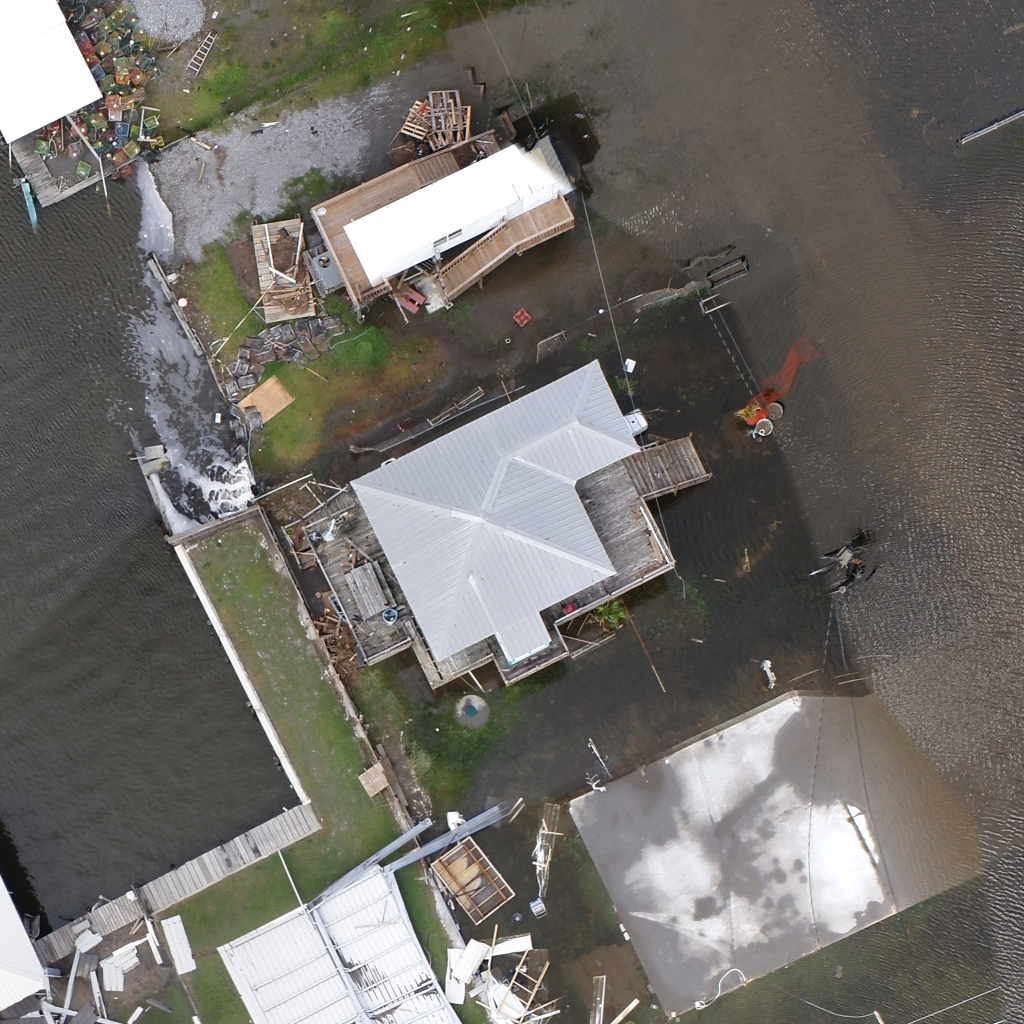} &
        \includegraphics[width=0.17\linewidth]{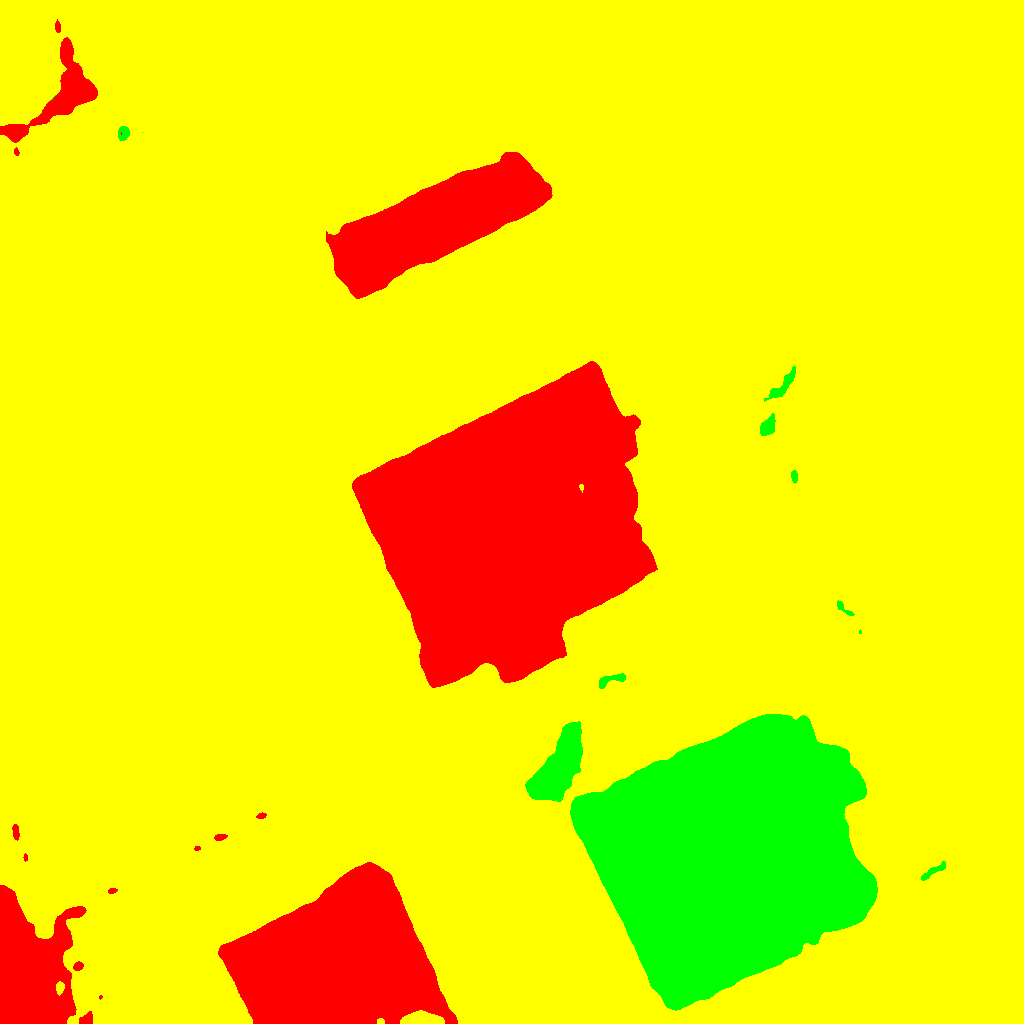} &
        \includegraphics[width=0.17\linewidth]{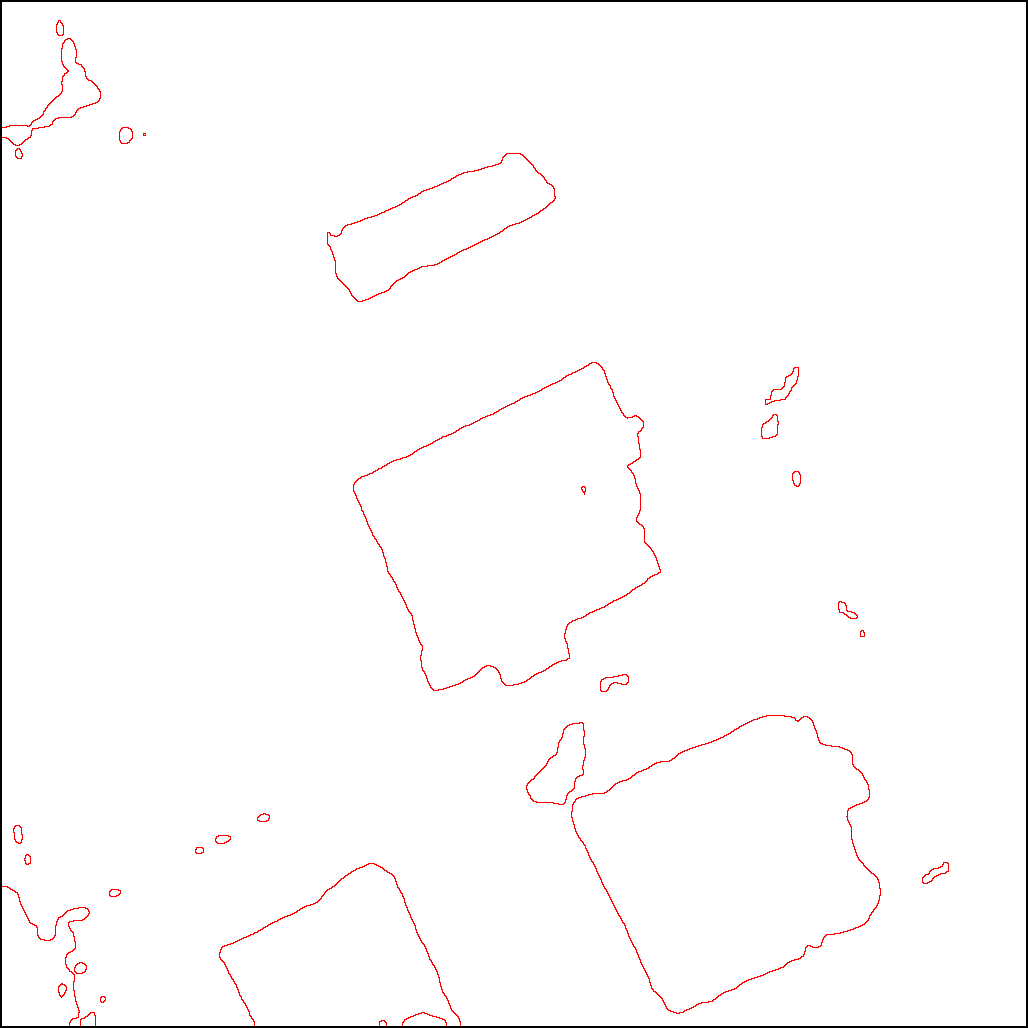} &
        \includegraphics[width=0.17\linewidth]{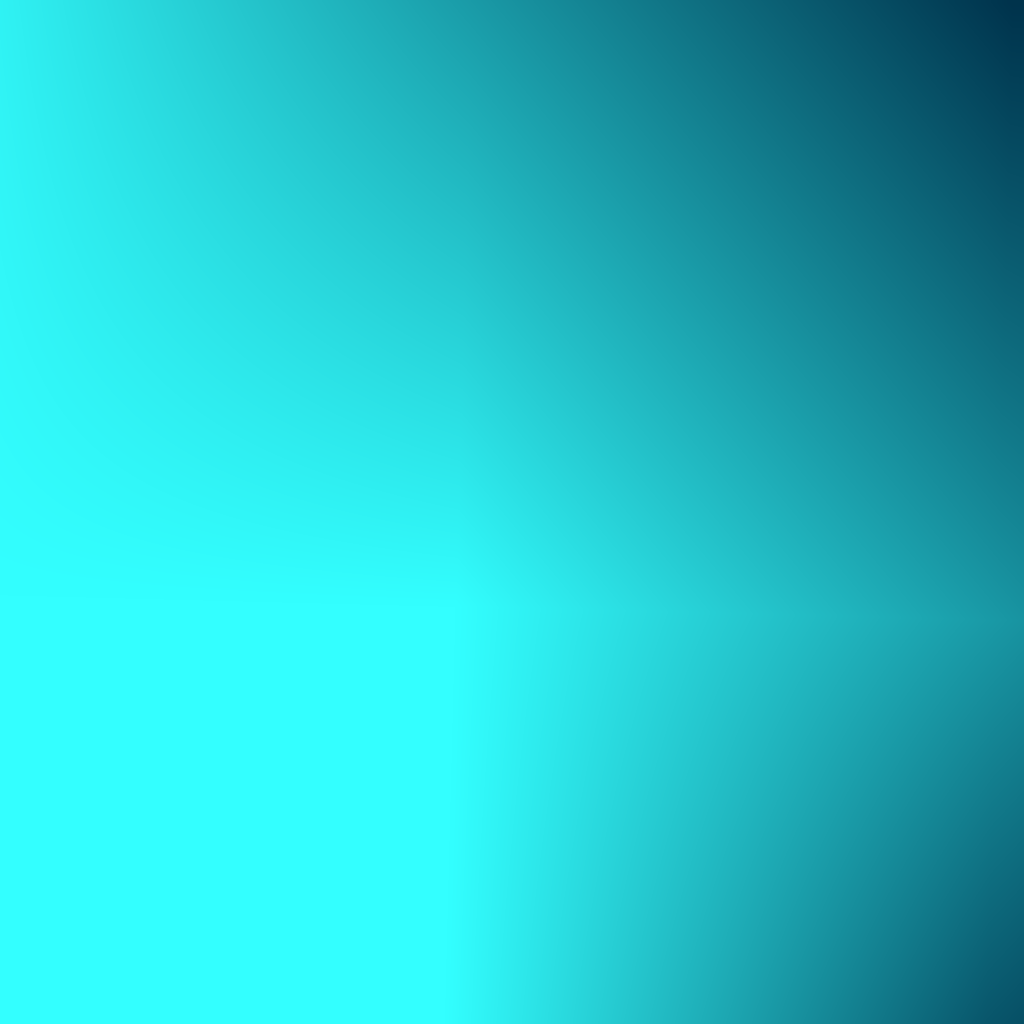} &
        \includegraphics[width=0.2\linewidth]{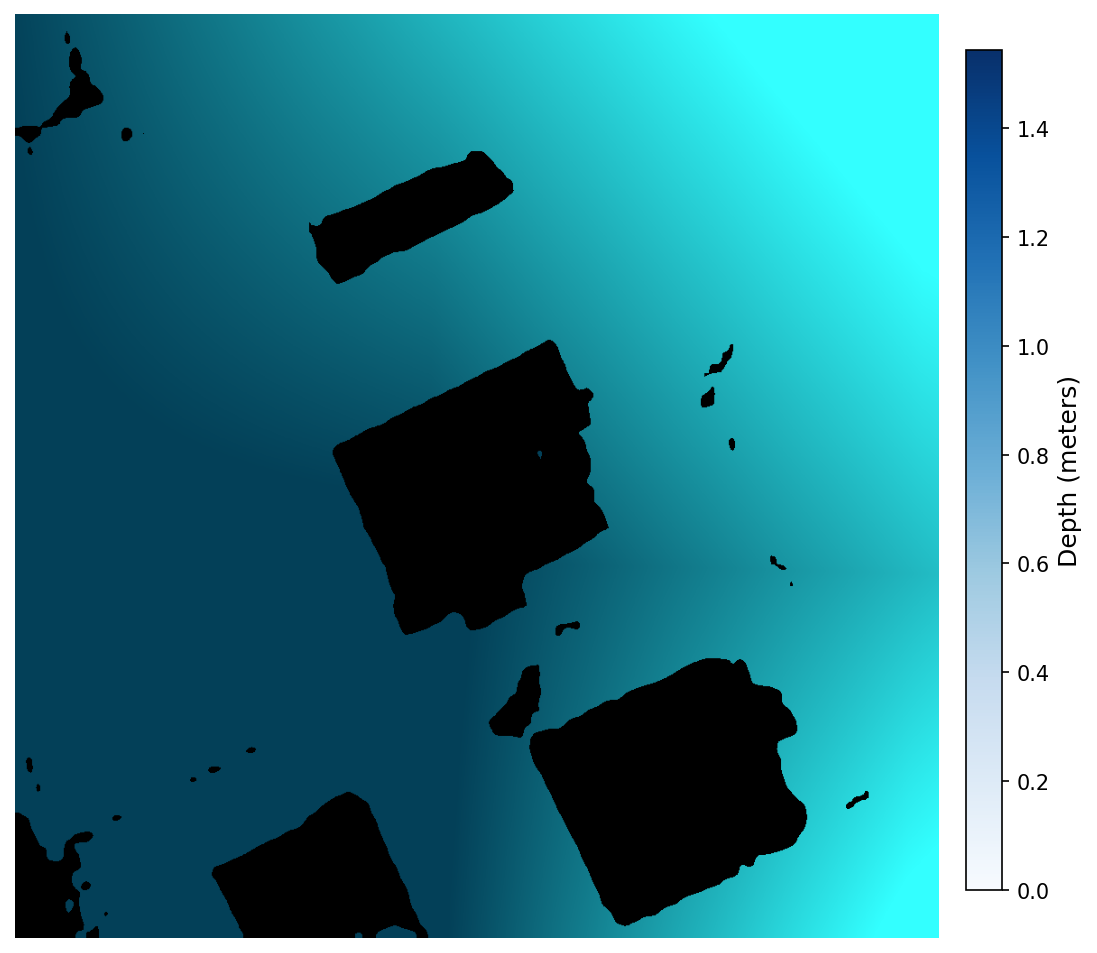} \\

    \end{tabular}
\caption{\textbf{Qualitative Results of Geometric Depth Estimation.} 
The figure illustrates our step-by-step pipeline for recovering flood depth: 
(1) \textbf{Input RGB}: Post-disaster aerial imagery; 
(2) \textbf{Segmentation Mask}: Semantic predictions distinguishing Flood (Yellow), Buildings (Red), and Background (Green); 
(3) \textbf{Water Boundary}: The extracted internal boundary (Orange) used to calculate water levels, demonstrating our ``Strict Dryland Contact'' logic; 
(4) \textbf{DEM Elevation}: The underlying Digital Elevation Model, interpolated to fill missing data; and 
(5) \textbf{Estimated Depth}: The final volumetric output. 
Both (4) and (5) share a consistent color depth visualization (Dark Blue to Light Blue) to facilitate direct comparison between the terrain topography and the computed water depth.}
\label{fig:qualitative_results}
\end{figure*}

(see Fig.~\ref{fig:qualitative_results}(3))
For each component $k$, we identify inner boundary pixels where water meets dry land via morphological erosion. Let $L^{(k)}$ be the binary mask of the $k$-th water component, and $S$ be a $3 \times 3$ structuring element (kernel). The erosion of $L^{(k)}$ by $S$, denoted as $L^{(k)} \ominus S$, is defined as the set of pixel locations $z$ such that the structuring element translated to $z$ fits entirely within the foreground of $L^{(k)}$:
\begin{equation}
    L^{(k)} \ominus S = \{ z \mid S_z \subseteq L^{(k)} \}
\end{equation}
where $S_z$ represents the translation of $S$ by vector $z$.
Boundary pixels at the image edge are discarded as field-of-view artifacts. Morphological erosion is adopted for its computational efficiency and parameter-free behavior, essential for rapid-response deployment.
\subsubsection{Step 3: Depth Calculation} (see Fig. \ref{fig:method_schematic})
To estimate the global water surface height, $Z_{water}^{(k)}$, for each flooded component, we aggregate the elevation values of all valid boundary pixels. Rather than using the maximum elevation - which is highly susceptible to segmentation noise, such as water masks bleeding onto vertical structures - we adopt a robust statistical approach. Following the methodology of \citet{cian2018flood}, we compute the 95th percentile of these boundary elevations. This threshold effectively filters out high-frequency outliers caused by misclassified vegetation or building walls, ensuring the derived water level represents a physically plausible upper bound for the standing water surface.
The depth for any pixel $(x,y)$ within component $k$ is then:
\begin{equation}
    Depth(x,y) = \max(0, Z_{water}^{(k)} - DEM(x,y))
\end{equation}
\subsection{Evaluation Metrics}
Depth accuracy is measured against FEMA flood depth rasters \cite{fema2017harvey}
using pixel-count-weighted averages across all tiles. For each tile $t$, let
$n_t$ denote the number of openly-flooded pixels in $\mathcal{F}_t$ (building
footprints and dry background excluded via the predicted flood mask), $\hat{d}_i$
the geometrically estimated depth, and $d_i$ the FEMA reference depth at pixel $i$.
The per-tile MSE is:
\begin{equation}
    \mathrm{MSE}_t = \frac{1}{n_t}\sum_{i \in \mathcal{F}_t}(\hat{d}_i - d_i)^2
\end{equation}
Aggregated across $T$ tiles weighted by $n_t$:
\begin{equation}
    \mathrm{MSE} = \frac{\sum_t n_t \cdot \mathrm{MSE}_t}{\sum_t n_t}, \qquad
    \overline{\mathrm{RMSE}} = \frac{\sum_t n_t \cdot \sqrt{\mathrm{MSE}_t}}{\sum_t n_t}
\end{equation}

\section{Experiments and Results}
\subsection{Experimental Setup}
Mask2Former \cite{cheng2021mask2former} backbone was fine-tuned on FloodNet \cite{rahnemoonfar2021floodnet} at $1024\times1024$ resolution on Jetstream2 \cite{jetstream,jetstream2}, and benchmarked against Attention U-Net \cite{oktay2018attention} and SegFormer \cite{xie2021segformer} on the FloodNet held-out test set (Table~\ref{tab:segmentation_results}). For volumetric evaluation, we use CRASAR-U-DROIDS \cite{manzini2024crasar} with aligned USGS elevation data \cite{krishnan2011opentopography}; since CRASAR lacks pixel-level water annotations, qualitative flood mask assessment is shown in Fig.~\ref{fig:qualitative_results}. For volumetric ground truth, we use FEMA flood depth rasters from Hurricane Harvey (2017) \cite{fema2017harvey}, derived from high-water mark surveys covering the Houston area. Since FEMA grids do not discriminate occluded structures, we apply the predicted flood mask to exclude building and background pixels before computing MSE and RMSE over openly-flooded pixels only.
\subsection{Segmentation Results}
\label{sec:experiments}
Table~\ref{tab:segmentation_results} reports performance on the FloodNet held-out test set. Mask2Former \cite{cheng2021mask2former} achieves 73.6\% mIoU, outperforming SegFormer (70.4\%) and Attention U-Net (69.7\%), with critical gains in ``Road Flooded'' (58.0\% vs.\ 51.4\%, 47.8\%) and ``Water'' (77.3\% vs.\ 73.6\%, 74.5\%) - the classes most responsible for waterline boundary precision in the depth pipeline.
\subsection{Quantitative Results}
Table~\ref{tab:depth_results} reports weighted MSE, $\sqrt{\mathrm{MSE}}$, and
 $\overline{\mathrm{RMSE}}$ on 136 Hurricane Harvey tiles from CRASAR-U-DROIDS
\cite{manzini2024crasar}, covering 52,890,333 openly-flooded pixels.
The pipeline achieves a $\overline{\mathrm{RMSE}}$ of 1.999\,m and a global
$\sqrt{\mathrm{MSE}}$ of 2.169\,m. The gap between the two reflects
tile-size heterogeneity: smaller tiles with lower individual errors pull
$\overline{\mathrm{RMSE}}$ below $\sqrt{\mathrm{MSE}}$, which is dominated
by the largest flooded tiles. Both values are physically reasonable given that
the method requires no hydrodynamic simulation or site-specific calibration.
\begin{table}[!t]
\centering
\caption{Flood depth accuracy vs.\ FEMA \cite{fema2017harvey} on 136 CRASAR
         Hurricane Harvey tiles (52,890,333 openly-flooded pixels).}
\label{tab:depth_results}
\begin{tabular}{|l|c|c|c|}
\hline
\textbf{Method} & \textbf{MSE (m$^2$)} 
                & \textbf{$\sqrt{\text{MSE}}$ (m)}
                & \textbf{$\overline{\text{RMSE}}$ (m)} \\
                \hline
Mask2Former & 4.702 & 2.169 & 1.999 \\
\hline
\end{tabular}
\end{table}
\subsection{Qualitative Results}
Fig.~\ref{fig:qualitative_results} validates the pipeline across three scenes.
Depth maps (Column~5) exhibit physically plausible hydrostatic gradients - shallow at the waterline, deepening toward basin centers. Three properties confirm geometric consistency: (1)~the ``Strict Dryland Contact'' algorithm isolates the true waterline (Orange, Column~3), rejecting image-edge artifacts;
(2)~Mask2Former correctly masks buildings (Red, Column~2), preventing erroneous depth inference on vertical structures; and (3)~depth gradients (Column~5) align with the underlying DEM topography (Column~4), confirming terrain continuity across all scenes.
\subsection{Sensitivity to DEM Accuracy}
Depth accuracy is inherently bounded by DEM quality. Vertical errors ($\sigma_{DEM}$) propagate directly into $Z_{water}^{(k)}$, and dense vegetation or urban canopy may reflect surface objects rather than bare-earth terrain, overestimating $Z_{water}$ and inflating depth. Future work will incorporate bare-earth DTMs and quantify this error propagation systematically.
\section{Conclusion}
We have presented a geometric framework for estimating flood depth by fusing transformer-based segmentation with topographic data. By replacing computationally expensive hydrodynamic simulations with direct observation, our pipeline delivers rapid, actionable volumetric insights from monocular aerial imagery. Qualitative validation, as highlighted in Fig.~\ref{fig:qualitative_results}, confirms the method's ability to recover realistic depth gradients in complex urban environments. Future work will focus on integrating this derived depth information into damage assessment networks to quantify structural damage levels for individual buildings.
\section{Acknowledgments}
This work was supported by NSF grants \#2423211 and \#2401942, the Consortium for Enhancing Resilience and Catastrophe Modeling, and NSF ACCESS allocation CIS251047 on Jetstream2 (NSF \#2138259, \#2138286, \#2138307, \#2137603, \#2138296).
\small
\bibliographystyle{IEEEtranN}
\bibliography{references}

\end{document}